# Generative artificial intelligence for *de novo* protein design

*Adam Winnifrith*[1], *Carlos Outeiral*[2], *Brian Hie*[3,4,5]

[1]*Department of Biochemistry, University of Oxford, South Parks Rd, Oxford, OX1 3QU, adam.winnifrith@keble.ox.ac.uk*

[2]*Department of Statistics, University of Oxford, 24-29 St Giles', Oxford OX1 3LB, United Kingdom, carlos.outeiral@stats.ox.ac.uk*

[3]*Department of Chemical Engineering, Stanford University, 443 Via Ortega, Shriram Center Room 129, Stanford, CA 94305, United States, brianhie@stanford.edu*

[4]*Stanford Data Science, 475 Via Ortega, Stanford CA 94305*

[5]*Arc Institute, 3181 Porter Dr, Palo Alto, CA*

**Abstract**

Engineering new molecules with desirable functions and properties has the potential to extend our ability to engineer proteins beyond what nature has so far evolved. Advances in the so-called "*de novo*" design problem have recently been brought forward by developments in artificial intelligence. Generative architectures, such as language models and diffusion processes, seem adept at generating novel, yet realistic proteins that display desirable properties and perform specified functions. State-of-the-art design protocols now achieve experimental success rates nearing 20%, thus widening the access to *de novo* designed proteins. Despite extensive progress, there are clear field-wide challenges, for example in determining the best *in silico* metrics to prioritise designs for experimental testing, and in designing proteins that can undergo large conformational changes or be regulated by post-translational modifications and other cellular processes. With an increase in the number of models being developed, this review provides a framework to understand how these tools fit into the overall process of *de novo* protein design. Throughout, we highlight the power of incorporating biochemical knowledge to improve performance and interpretability.

## Introduction

Proteins are highly complex nanomachines that perform a range of functions in biological systems, mediating intricate regulatory networks, catalysing reactions at ambient temperatures, and forming materials with exceptional tensile strengths. Inspired by the variety of chemical functionalities displayed by natural proteins, designing artificial proteins for specific applications has been a long-pursued objective in synthetic biology. Experimental methods such as phage display [1] and directed evolution [2] have been developed to identify or enhance proteins with desirable functions; however, they are limited by size and cost, and are constrained to use natural proteins as starting points. On the other hand, computational strategies have been developed as alternatives to efficiently generate protein sequences with desired functionalities. Over recent years, there has been a rapid acceleration in the development of computational protein design, to the extent that the most sophisticated approaches now vie with experimental screening techniques. The emergence and subsequent proliferation of these methods forms the central theme of this review.

# Introduction to *in silico* protein design

## Protein space

*De novo* protein design is often conceptualised through the framework of protein space, an abstract landscape where every coordinate represents a protein sequence, with an additional coordinate signifying the protein's *fitness* or ability to perform a function (Fig. 1). Navigating this vast protein landscape is the primary challenge of protein design, and the multitude of peaks and valleys that can trap a search process within suboptimal designs (the so-called *ruggedness* of the landscape) represent a traditional obstacle to generate novel proteins.

Traditional *de novo* protein design utilises physics-based methodologies with iterative search strategies, aiming to navigate protein space through modelling the biophysical interactions governing the folding, shape, and function of designed sequences [3]. Tools, like Rosetta, developed based on these approaches, have been successfully employed over the past 20 years to engineer numerous proteins *in silico* [4,5]. An insightful review and visual timeline of de novo designed proteins using these approaches is provide by Woolfson [6]. However, these models are computationally expensive, often have to make assumptions that limit their predictive properties, and are prone to stalling at local optima, hindering the discovery of diverse functional variants. Ultimately, understanding the intricacies of protein physics remains one of the central problems in modern biochemistry, and as physics-based designs methods near the boundaries of our understanding, their capabilities become more and more limited.

## Generative models

An alternative approach to physics-guided protein design involves selecting candidates with desired properties exclusively from 'plausible' regions of protein space, signifying proteins that have a reasonable likelihood of performing specific functions of interest. This search is often enacted by building (or *training*) a generative model — a statistical model that learns a distribution over protein space. As the model can simultaneously model the variables that determine protein sequence and fitness, it is then possible to "sample" protein instances from regions of high fitness for a particular task.

The advent of deep learning, which enables learning vastly complex probability distributions, has enabled a plethora of generative models. In this review, we organise the latest *ab initio* design techniques into two groups: sequence-based and structure-based approaches. Structure-based design has recently focused on the use of *diffusion models* (the models behind text-to-image generators [7–12]), although with a significant contribution from *graph neural networks* (GNNs), whereas sequence-based design has primarily focused on *large language models* (LLMs) (the models behind ChatGPT [13–16]). As we will argue, these models have learned to generalise beyond the parts of protein space inhabited by natural proteins, and can venture areas of this space unexplored by nature [17,18]. Moreover, they are enabling scientists to specify desired properties using easy-to-use programming languages [19] or even natural language text [20,21]. Recent advances in improving *in silico* checks might further increase the success rate of designed proteins to close to 100% [22].

## Protein structure prediction

Protein structure prediction algorithms (which we term sequence-to-structure algorithms, Fig 3.), such as AlphaFold (AF) [23], RoseTTAFold (RF) [24,25], RoesTTAFoldAA (RFAA) [26], ESMFold [27], OmegaFold [28], OpenFold [29] or UniFold [30] have particularly benefited *de novo* protein design for two reasons. First, they have provided *in silico* metrics to score predicted designs. Bennett et al [18] discovered that filtering designs based on the metrics *predicted Local Distance Difference Test* (pLDDT – a per-residue model confidence measure) and *Root Mean Squared Deviation* (RMSD) between the coordinates of atoms in the structures from the generative model and coordinates predicted by a structure prediction algorithm on the sequence of the designed protein, increased experimental success rates tenfold. Secondly, they have provided an abundance of structural training data for other models. For example, Hsu [31] used 12 million protein structures predicted by AlphaFold2 to train the structure-to-sequence model ESM-IF.

However, one concern with their use in protein design is their adaptability to novel topologies and folds absent from training sets. These methods also exhibit limitations in predicting physiologically relevant protein states or components. For instance, they cannot yet model protein conformational landscapes [32], and whilst RFAA can now model ligands, co-factors and post-translational modifications (PTMs), knowledge of the which combination of PTMs are present under different conditions still requires experimental methods to determine. Structure prediction methods are based, by and large, on deep neural networks trained on large corpuses of crystal structures, and as such they may be considered akin to structural studies of proteins, which while deeply insightful are also limited to specific states of the proteins that may not well represent how they behave in the cell.

## Sequence-based protein design

We commence our discussion of protein design with sequence generation (properties-to-sequence algorithms, Fig 3.) which primarily uses *large language models* (LLMs) to produce novel protein sequences with similar properties to natural proteins. The way LLMs have addressed the challenge of sampling from plausible protein space can be thought of as learning the grammar of protein language (see Fig. 3a and Box 1). Models are trained to predict the identity of masked amino acids when given the context of some parts of a protein sequence (with the exact parts depending on the model type) [33]: during training, the model's predictions are compared to the known sequence and its weights are then updated based on whether it made a correct or incorrect prediction. When asked to generate new proteins, models typically predict amino acids sequentially to build up a novel protein sequence. Increasingly, these sequences have both low homologies to known proteins, similar physicochemical properties to natural proteins [34], and high experimental success rates suggesting an ability to sample from plausible protein space unexplored by nature.

One of the first examples of protein design using language models was Verkuil *et al*'s model [17] which generated proteins incorporating complex structural motifs, such as hydrogen bond networks. Of the 152 successful experimental designs (out of 228 tested) in Verkuil's models, 35 had no significant sequence match to known natural proteins, and the remaining 117 had a median sequence identity of 27% to their nearest matches. **ProGEN** [35] also demonstrated this learning by achieving an experimental success rate on a set of generated proteins comparable to controls set of random and natural proteins. Furthermore, Verkuil's model, **ProtGPT2** [34], and **ZymCTRL** [36] also demonstrated low *structural* identity to known proteins.

Large language models can also generate protein sequences that correspond to a specific function by utilising conditioning tags or Markov Chain Monte Carlo (MCMC) searches (see Box 1). For example, during the training of ProGEN and ZymCTRL, enzyme database classification codes were attached as tokens to the start of the protein sequence. This enabled the models to successfully learn different probability distributions for different enzyme families. When ProGEN was used to generate proteins with 'lysozyme' activity, 66 out of the 90 designs tested experimentally were functional.

Language models have also played a leading role in the parallel role of protein engineering, where the objective is, instead of designing a protein completely *de novo*, to optimize some properties of a known protein. The ability of language models to learn the grammar of protein space also means that they are excellent starting points to build models that can predict a variety of properties (which we denote as seq-to-property models, Fig 3.). The intermediate states in the language models (known as *latent variables* or, more commonly, *learned representations*) can be used as rich inputs for a variety of machine learning models. Representations from language models are central to the state-of-the-art tools for predicting protein function [37–40], subcellular localisation [41], solubility [42], ligand (drug or substrate) binding sites [43–45], signal peptides [46], some post-translational modifications [47–49], and intrinsic disorder [50], among others. In turn, these models can be used in sophisticated pipelines to optimise the properties of a starting protein. We refer the reader to reviews by Hie *et al*. [51] and Yang *et al.*[52] for the application of these models to protein engineering.

Despite these early successes, limitations remain. While generating novel topologies that perform well-defined functions using the ample examples across the tree of life has been successful, new structures with novel catalytic mechanisms or entirely new functions are yet to be created using language models, despite this having been previously achieved before the advent of deep learning [5,53,54]. Furthermore, search methods like MCMC for obtaining desired properties or functionality, or to optimise protein properties, can be time-consuming due to the extensive number of iterative generations required and their tendency to become trapped in local minima. This issue is particularly pronounced when exploring regions of space where the predictive capabilities of the LLM struggles to extrapolate effectively.

To overcome some of these limitations, an alternative sequence-based design methodology has been to utilise text-based diffusion models [55,56]. In the next section we discuss diffusion models' performance on design tasks that leverage structure-based information, however they can also be set up to use series of tokens as an input in a similar way to LLMs. Briefly, diffusion models that operate on text tokens can be thought of as a language model that learns to generate sequences through all possible decoding orders rather than solely left-to-right. The first sequence diffusion model to be trained on the entire UniProt sequence space was Evo-Diff [57], which leveraged its diffusion objective to generate intrinsically disordered regions of proteins – a task intractable for structure-based techniques due to the impossibility of obtaining single structures for IDRs. Protein Generator [58] and DiffSDS [59] are also sequence based diffusion models, however they are trained using structurally information as well so we discuss them in the next section. They highlight how newer model architectures that can incorporate both sequence and structural information are starting to disrupt the classical distinction between the two design paradigms.

## Structure-based design

Structure-based design methods, in contrast to sequence-based methods, focus on creating novel protein structures (properties-to-structure algorithms, Fig. 3.). We will focus our discussion on diffusion models, a recently introduced type of deep generative model that has shown promise for full protein structure generation. However, we will first briefly mention two other structure-based techniques, hallucination and reinforcement learning, and discuss their limitations in generative modelling. We will then introduce inverse folding (structure-to-sequence models, Fig 3.), as an essential technique for structure-based design that enables sequence generation from structural models that has seen significant recent advancements.

Prior to the introduction of diffusion models, the advent of rapid and accurate deep-learning based protein structure prediction algorithms enabled structure-based design methods like *hallucination* [60,61] (reviewed in [62]), where the structure predictor is subject to a search algorithm, typically using MCMC, to find sequences that fold onto the correct shape. While this technique is subject to ongoing improvements [63], it doesn't generate a probability distribution over protein space, it is computationally slow, requiring a large number of evaluations, and it is limited to designing a scaffold around a number of well-defined sites. A similar strategy to iteratively sample plausible proteins was recently introduced that used reinforcement learning in combination with a Monte Carlo Tree Search to iteratively design proteins in a “top-down” approach [64]. Whilst very successful in designing nanopores and ultracompact icosahedra, it also is not truly generative, in the sense that it does not model a joint probability distribution over protein coordinates and fitness, and therefore has limitations in generating novel proteins.

## Inverse folding

Generating sequences that fold into a given structure — termed *inverse folding* by the community and which we refer to as structure-to-sequence algorithms — constitutes a crucial aspect of numerous structure-based *de novo* protein design methodologies [3]. Initial approaches to inverse folding relied on physics-based approaches, like Rosetta, to generate sequences. Recently, graph-based deep learning techniques (also known as *geometric deep learning*) have proven remarkably adept at this task [31,65–75]. Graph neural networks conceptualise each backbone atom of protein’s structure as a node on a graph, and connections are drawn to all other atoms within a defined distance, forming the graph's edges. Within the model's architecture, information is passed between these proximate atoms such that each node contains information about its surroundings, creating a structurally-informed representation of the protein. The success of graph neural networks at inverse folding is exemplified by their use to redesign previously unsuccessful protein structures, increasing experimental success rates [71]. Tools like ProteinMPNN have repeatedly improved the yield of soluble protein from initially unsuccessful designs [76–79]. Several other graph-based inverse folding techniques exist, including ESM-IF [31], PiFold [73], MIF-ST [72], and Knowledge-Design [75].

The ability to produce sequences that fold to a given protein structure is not limited to single structures. For example, the generative process can be biased to generate protein assemblies with desired stoichiometries and symmetries [18,60].

## Diffusion models

The success of inverse folding methods in protein design has been curtailed by the limitation of having an initial protein structure that can be successfully translated into a sequence. In the past year, significant interest has focused on *diffusion models*, a type of generative deep learning architecture that sidesteps this limitation. The way diffusion models [7–9] have tackled the challenge of sampling from plausible protein space can be thought of as learning the process of drawing proteins (see Fig. 3b and Box 2). During model training, protein structures that have been determined by structural biology methods such as X-ray crystallography and CryoEM are noised by iteratively adding small amounts of Gaussian noise. The model is then trained to recover the original protein structure from the noised data. Specifically, a neural network is employed to predict either the completely denoised protein structure or gradient of the deterministic reverse of the noising process from noised structures with varying amounts of noise added over many different protein structures (see Box 2 for details). A loss function on the difference between the true denoised protein structure or true gradient at that noise level is then used to train and update the weights of the neural network. This enables the model to learn probability distributions of protein space such that, at inference time, the model can be provided with true Gaussian noise and iteratively generate novel protein structures.

Initial remarkable success in sampling from plausible protein space beyond that explored by nature was achieved by diffusion models that generated backbone coordinates to produce diverse monomers and protein assemblies [80]. Proteins generated by **Chroma** [20] contained a similar distribution of fragments of protein structures — reused by unrelated proteins — to natural proteins. **RFDiffusion** [18] generated CryoEM-validated novel alpha-beta barrels architectures that had 16:16 and 12:24 helix:strand ratios – extending beyond the structural variations of the classic 8-strand/8-helix TIM barrel fold explored by evolution. These models are backbone-only (see next section for further improvements) and require a final inverse folding step where one of the methods in the previous section is used to map the structural model to a protein sequence.

Diffusion models can also be biased to incorporate desired properties and functions throughout the generation process (see Box 2, Fig D), similar to other generative models like LLMs. The conditioning process results in quicker, more effective outcomes and enabling more semantic conditioning. For example, **RFDiffusion** was employed to generate de novo binders for five different targets, achieving an 18% success rate with 95 designs, and created binders to helical peptides with picomolar specificity [81]. **Chroma** and **ProteinDT** [21] can be further conditioned on natural language annotations. This introduces the tantalising potential for protein design queries to be articulated in natural language — for example, "Design a protein that binds to PD-L1, exhibits solubility, and exists as a monomer."

## Improvements to diffusion models

An important recent development has been how to represent proteins to ensure output invariance to rigid transformations of the input data, such as rotation and translation. For example, RFDiffusion employs a frame representation for each residue comprising its Cα coordinate and N-Cα-C rigid orientation. To noise the inputs during training, 3D Gaussian noise is applied to the Cα coordinates and Brownian motion is applied to the residue's orientation. RFDiffusion leverages RoseTTAFold's pretrained weights and ability to directly predict protein structures using a loss function on the difference between the true and predicted denoised structures (see box 2, fig C). Other methods have been recently developed to create a diffusion process directly on SE(3) data distributions. **Genie** [82] performs diffusion on Frenet-Serret frames (representations of the position and orientation of residues relative to a global reference frame) whilst **FrameDiff** [83] performs diffusions on the rotations and translations (3D coordinates) separately, as in RFDiffusion, but uses an independent score network for each. This allows FrameDiff to operate using a loss of the difference between the true and predicted gradients of the denoising process (see box 2). FrameDiff achieves similar performance to RFDiffusion with a quarter of the parameters – an important consideration for computational protein design where generative models are often run tens of thousands of times and the cost of running the model each time scales with its size. This advance highlights the importance of the advances in developing the SE(3)-equivariant representations of proteins that are compatible with deep learning architectures and that capture the symmetries of 3D space.

Both **Protein SGM** [84] and **FoldingDiff** [85] were earlier alternative diffusion approaches. ProteinSGM represents proteins as a series of matrices and recovered their structure using Rosetta whilst FoldingDiff models the backbone as a series of angles rather than 3D coordinates. This angle-based approach, however, is subject to 'lever' effects which refer to the phenomenon where small mistakes introduced at early stages of a process (in this case generating the backbone protein angles) can lead to large-scale alterations in the final outcome (protein structure). Consequently, RFDiffusion and FrameDiff have outperformed both Protein SGM and FoldingDiff in designability. Despite this, the principle employed in FoldingDiff of elegantly incorporating more biochemical information by more closely mirroring the biophysical process protein folding could yield successful outcomes if the 'lever' effects can be effectively controlled, potentially requiring the use of larger models.

Diffusion models' functional generation has largely been confined to binder design and scaffolding without venturing extensively into designing proteins with novel enzymatic activity. A crucial element of enzyme design centres on finding the right arrangement of side chains around the binding site to stabilise the transition state. The state of protein diffusion models in Spring 2023, however, required sequential generation of backbones, sequences, and rotamers by structure-to-sequence and rotamer packing modules. A major limitation that means ligand-side chain and backbone-side chain interactions are largely ignored. However, newer generation diffusion models, especially those that incorporate side chains, offer significant promise in fine-tuning side chain positions for transition state stability.

One such all-atom diffusion model was recently proposed by Chu and colleagues [86], which simultaneously designs sequence and structure. In their model, named **Protpardelle**, an all-atom protein structure is predicted at each time step and then the backbone coordinates of this model are used to predict a sequence using a modified structure-to-sequence module. The predicted sequence is subsequently fed into the next time step and all the atoms of the side chains of that sequence are diffused, with a diffusion size proportional to the last time the atoms appeared in the diffusion process. This allows the calculation of an updated final structure prediction encompassing all atoms but allows the sequence to change with each time step by utilising the backbone to propose a new sequence.

A related approach to Protpardelle is **RFDiffusionAA** [26] which is not only a protein all-atom model but also models all the atoms of ligands and post-translational modifications. This was achieved by first developing a modified RoseTTAFold that was trained on information about the chemical structure, bonding, and chirality of small molecules and PTMs bound to proteins. As with RFDiffusion, a diffusion model was then trained using the network weights of this model. In contrast to Protpardelle the model does not generate every atom of the protein at every step in the diffusion process, rather a limited selection of residues from the protein are atomised into their atomic constituents. RFDiffusionAA was used to generate binders against digoxigenin (0.07% success rate), the heme cofactor (57% success rate), and the bilin pigments (3% success rate) involved in the light accumulation for photosynthesis. All the proteins designed had low sequence and structural similarity (<0.62 TM score) to known proteins.

As we highlighted previously, parallel progress has been observed by performing diffusion on protein sequence space. Here we focus on sequence diffusion models trained whilst incorporating structural information. In **Protein Generator** [58] diffusion is performed by initialising each amino acid position as a vector of 20 units in length, with each unit assuming a value between 1 and -1 and each digit corresponding to a specific amino acid (a technique commonly used and referred to as one-hot encoding). In a given protein sequence, the true amino acid assumes a value of 1 while all others are denoted as -1. Gaussian noise is subsequently applied to these units. Similar to the methodology of RFDiffusion, the authors employ RoseTTAFold as the denoising neural network, made possible by RoseTTAFold's capability to predict both sequence and structure from noised protein sequence data, a feature resulting from sequence masking during training. The loss function for Protein Generator is based on both the predicted sequence and structure. A distinctive aspect of the training protocol was the inclusion of additional tasks, namely sequence-to-structure and structure-to-sequence. Performance levels on par with RFDiffusion were achieved for proteins with up to 300 amino acids in length but declined beyond this. However, as diffusion takes place on sequence space, sequence classifiers can be integrated into the model during each denoising step, thus guiding the trajectory towards desired outcomes. This approach allowed the design of proteins with specified amino acid compositions, similarities to known families or folds, specific motifs scaffolded, and proteins capable of switching folds between two states when cleaved, as discussed later.

As with the application of diffusion models to sequence-based design, there have been attempts to incorporate more structural information into language models. For example, **SaProt** [87] leveraged the structural alphabet of **FoldSeek** [88] (an alphabet that provides each residue with one of 20 letters to describe its geometric conformation with its spatially cloest residue) to create a protein language model that outperformed all previous language models on being able to predict properties of proteins. Although not used as a generative model in this case, the performance of SaProt highlights the potential the incorporating more structural information into language models could provide.

## Designing dynamic proteins

A critical feature of proteins, enabling their diverse functional range, is their ability to change shape rapidly and dynamically. For example, membrane transporters alternate between open and closed states, with gating facilitated by voltage-sensing domains. Bacteriocins display impressive conformational changes as they thread dynamically through pores [89]. Progress has been made in structure-to-conformation models through machine learning predictions of molecular dynamics simulations, employing both convolutional neural networks [90] and diffusion models [91]. Initial sequence-to-conformer models include manipulating the MSA [92,93] and constructing diffusion models based on protein representations as a series of springs instead of coordinates [32]. However, these current conformational modelling approaches have achieved limited success in efficiently sampling a protein's energy landscape. A key benchmark in future work should involve testing whether these models incorrectly predict multiple conformers for proteins that do not possess them.

Initial success has also been achieved in designing conformations-to-structure workflows. Praetorius et al [94] designed a new version of the eight alpha helical bundle designed-helical-repeat protein that could switch between two states, with one stabilised by a binding peptide obtained. This was achieved by *in silico* designing a new open conformation of the protein through combining two rotations of the protein and splitting the two rotations into three different domains (two four helical bundles and a single helix to act as a binding peptide). Achieving an experimentally successful protein required extensive use of physics-based Rosetta pipelines for interface redesign and energy calculations in combination with the deep learning tools ProteinMPNN and AF2 to generate sequences and rank designs. Impressively, ProteinMPNN was used to collect consensus sequences that produced mutations that stabilised each state, allowing the equilibrium thermodynamics to be modulated.

Protein Generator [58] was used to design proteins that change between different conformational states upon cleavage. One example of this was scaffolding the pore forming protein melittin. The authors provided the melittin sequence and a protease cleave site to the model, which was then tasked with generating an additional 125 residues and necessitating the protease cleavage site to be in a loop. From the 12 designs selected for synthesis, 9 exhibited the correct secondary structure when expressed as monomers.

## Discussion and future perspectives

Generative models' ability to sample from plausible protein space has enabled dramatic advanced across an exciting set of *de novo* protein design problems. However, the breadth of successes has only highlighted the obstacles for protein designers that remain. We dedicate the last portion of this review to considering these future perspectives.

We propose understanding protein design through a framework highlighting five aspects of proteins important for protein design (Fig. 3). What sequence of amino acids are they comprised of? What three-dimensional structure do they fold into? How can this structure dynamically change shape? What does this protein structure do and what physiochemical properties does it have? And, finally, how is the protein's function regulated? The field of computational protein design has developed tools to connect these pieces of information: for example, protein structure prediction algorithms convert a 1D sequence into a 3D structure (sequence-to-structure), whereas inverse folding algorithms perform the opposite step (structure-to-sequence).

Key gaps highlighted by this framework include the inability to generate conformational landscapes of proteins based solely on sequence or structure without the aid of costly molecular dynamics simulations, limited capacity to design proteins that undergo conformational changes, and incorporating structural regulation such as a phosphorylation on-off switch. Addressing these intricacies could pave the way for creating proteins responsive to small molecule ligands or for generating novel enzyme mechanisms. Specifically, devising these unique enzyme mechanisms might involve sampling multiple conformations that stabilise a transition state while maintaining a low affinity for the product.

Generative deep learning is currently limited by its ability to understand the physical interactions underpinning molecular behaviour. For example, current approaches, which are trained on canonical amino acids, could encounter difficulties when ambitiously expanding de novo design to include non-canonical amino acids, even though some may present very similar physicochemical properties to canonical ones. One might consider training design algorithms on the structural data of proteins produced using structure prediction algorithms. However, these structure prediction algorithms often fall short in capturing the underlying physics of protein folding [95]. More critically, they have so far tended to inadequately predict the impact of post-translational modifications (PTMs) on structures, though RFAA looks set to changes this. Water molecules are also often overlooked in protein design, despite being very important for protein interactions, in particular for TCR interactions [96] and membrane pores [97]. These facts suggest that a central factor towards improving the performance of deep learning models in de novo design could be to introduce effective physics-inspired inductive biases in the deep learning architectures.

Incorporating biochemical knowledge has been a key theme in developing improved generative models for protein design and engineering. Several studies have shown that incorporating biochemical insights in the training data, data processing and protein representation leads to significantly improved outcomes across different tasks. This was exemplified by Johnson [22], when removal of signal peptides in their data pre-processing resulted in non-functional proteins, and by Outeiral [98] who developed a language model trained on codons that outperformed larger protein LLMs. Furthermore, it is important to emphasise that these data-driven approaches are underpinned by large, open-source databases curated through decades of biochemical research, demonstrating the power of collaborative open science.

High-throughput, information-rich protein production and biochemical assays would further accelerate the development of the field by enabling rapid testing of hundreds of designs for desired functions. For enzymes, a proof-of-concept step towards this was recently provided by through the use of a cloud lab to conduct high throughput DNA assembly, protein-expression, and functional testing – dubbed Self-driving Autonomous Machines for Protein Landscape Exploration (SAMPLE) [99]. High-throughput screening of antibody-antigen interactions, whilst still obtaining sequence-to-function data has been achieved by conducting transcription, translation and screening of antibody libraries on an Illumina flow cell [100]. Robotics enabled automated protein fitness landscape exploration tools combined with the development of diffusion models on sequence space (as with ProteinGenerator) open up the opportunity for iterative rounds of active learning without scientist input. These tools could help overcome current challenges by developing experimentally validated *in silico* metrics to increase designs success rates, accurate classifiers of protein function to condition diffusion models, and data for reinforcement learning.

Novel advances in deep learning are also likely to permeate into de novo protein design. As the recent development of diffusion models rapidly resulted in the tools outlined above, it is likely that as more advanced models are developed in other areas, for example the recent development of consistency models [101], PGFM++ [102], or Geometric Neural Diffusion Processes [103] as alternatives to diffusion models, they will be translated into improvements for protein design. An alternative approach might be the combination of models - combinations of LLMs and diffusion models have already been developed [21,59].

The developments outlined in this review also highlight that protein design is becoming increasingly accessible to a broader spectrum of research labs. This phenomenon is perhaps best exemplified by the release of the RFDiffusion code and its accompanying interactive control suite. This democratisation promises to usher in revolutionary advancements, offering transformative solutions to pressing global challenges ranging from healthcare breakthroughs to environmental remediation. It is likely that the initial breakthroughs will be seen in sectors such as manufacturing, plastic degradation, and carbon capture — areas where the barriers to market entry are lower due to the absence of lengthy clinical trials. Yet, even as these innovations take centre stage, the broader potential for healthcare and other sectors remains profound. However, as the field progresses, it is imperative that researchers remain aware of potential biosecurity risks. A judicious and proactive approach will be vital to ensure that the advancements in de novo protein design are harnessed safely.

We envisage protein design evolving not only into a staple research tool for developing binders and control proteins but also as a key instrument to address and resolve grand challenges, ultimately bettering our world.

## References and Recommended Reading

Papers of particular interest, published within the period of review, have been highlighted as:
• of special interest
•• of outstanding interest

• The authors integrate protein LLMs, sequence-to-structure and for fixed backbone design, using MCMC sampling of a conditional distribution based on the desired shape, and unconstrained generation, using a blocked Gibbs sampling approach. The model successfully generalised beyond known protein space, and the designed proteins were validated through experimental evaluation.

•• Diffusion model utilising RoseTTAFold-based protein structure prediction network as the denoising component. The authors highlighted two important aspects of their architecture that were crucial for the model's performance on *in silico* success rates: 1) inspired by recycling in AlphaFold, self-conditioning the model on the predicted final structure of the previous step 2) utilising Mean Squared Error loss during model training, likely because it is not invariant to the global reference frame and therefore promotes continuity between timesteps. The authors use the model to generate proteins conditioned on shapes, scaffold enzyme functional sites, and generate *de novo* binders. Extensive experimental validation with structural and functional data demonstrated success rates of

~20% across different design tasks from sets of just 96 expressed proteins. Of particular utility to the community is the open-sourced easy-to-use code.

• The authors introduced several innovations that created efficient, scalable diffusion models for proteins. These included novel SE(3) equivariant representations of proteins using random graph neural networks built up from SE(3)-invariant features such as interatomic residues and incorporation of long-range interactions into these graphs with sub-quadratic scaling using N-body simulation methods. Several classifier models trained on noised structures were used to program shape, symmetry, scaffolding, secondary structure, and natural language annotations.

• Experimental evaluation of the accuracy of sequence-based, structure-based, and homology-based metrics in predicting the functionality of CuSOD enzyme designs generated by three distinct generative models. Notably, the widely used pLDDT metric was not indicative of experimental functionality, with the ProteinMPNN and ESM-1v log-likelihood scores better explaining activity. The authors provide a helpful learning example for readers of the importance of biochemically informed data pre-processing. Initially signal peptides and transmembrane domains were truncated, which resulted in many non-functional proteins from the generative models. Analysis of the protein crystal structure revealed CuSODs are active as homodimers with the dimerisation interface in the truncated regions. Accounting for this improved the models' success rate.

## Acknowledgements

The authors are grateful for the many helpful discussions that we had with Piotr Jedryszek, Weronika Slesak, Phil Biggin, Charlotte Deane, Alissa Hummer, Tobias Olsen, George Wicks, and Nathan Ewer who provided incredible feedback and guidance. Carlos Outeiral thanks Schmidt Futures for support through an Eric and Wendy Schmidt *AI in Science* Postdoctoral Fellowship.

**Figure 1**

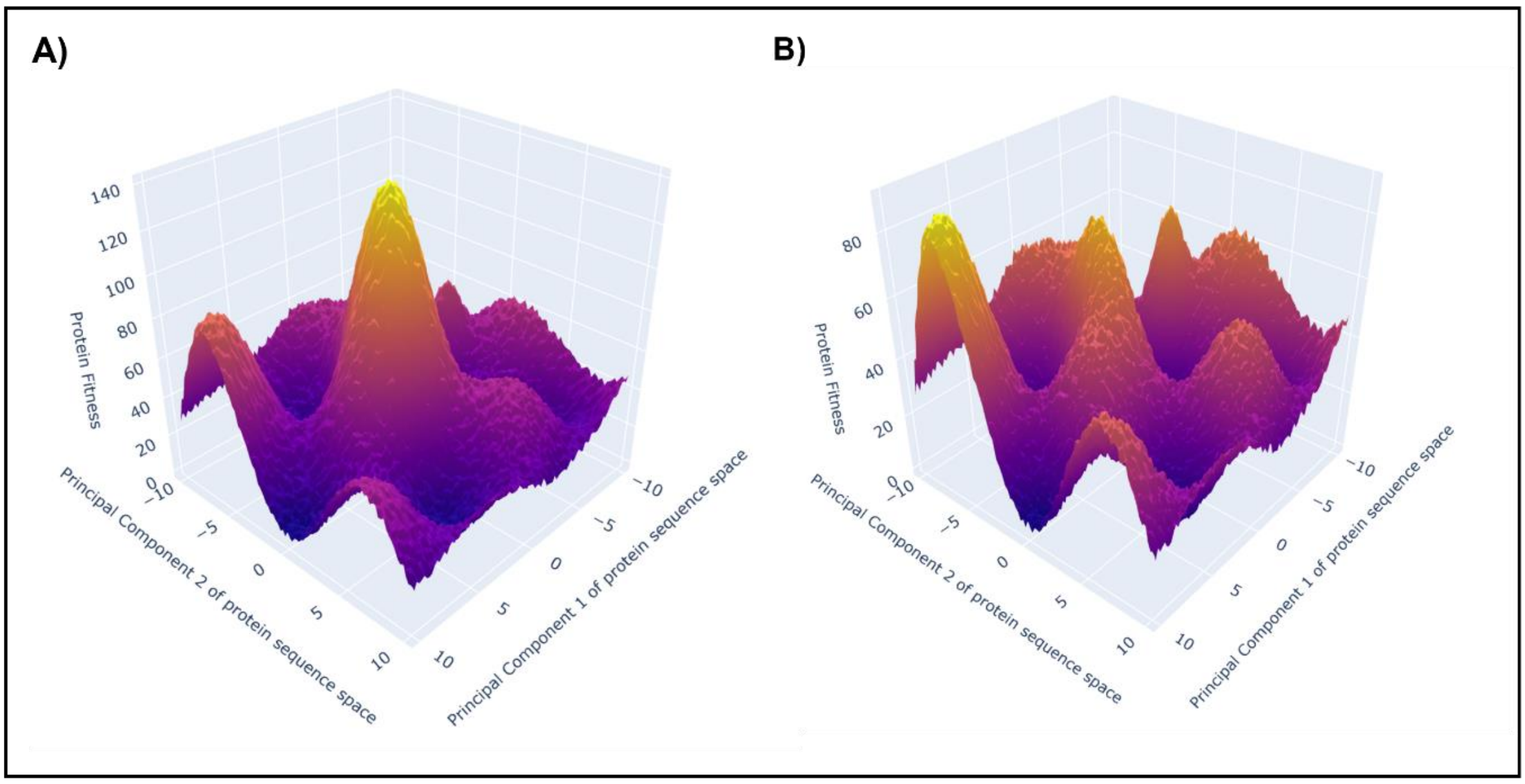

Example protein fitness landscapes **a)** This protein fitness landscape has a clear local maxima with high fitness **b)** A more rugged protein fitness landscape with many maxima and minima that make it hard to navigate

**Figure 2**

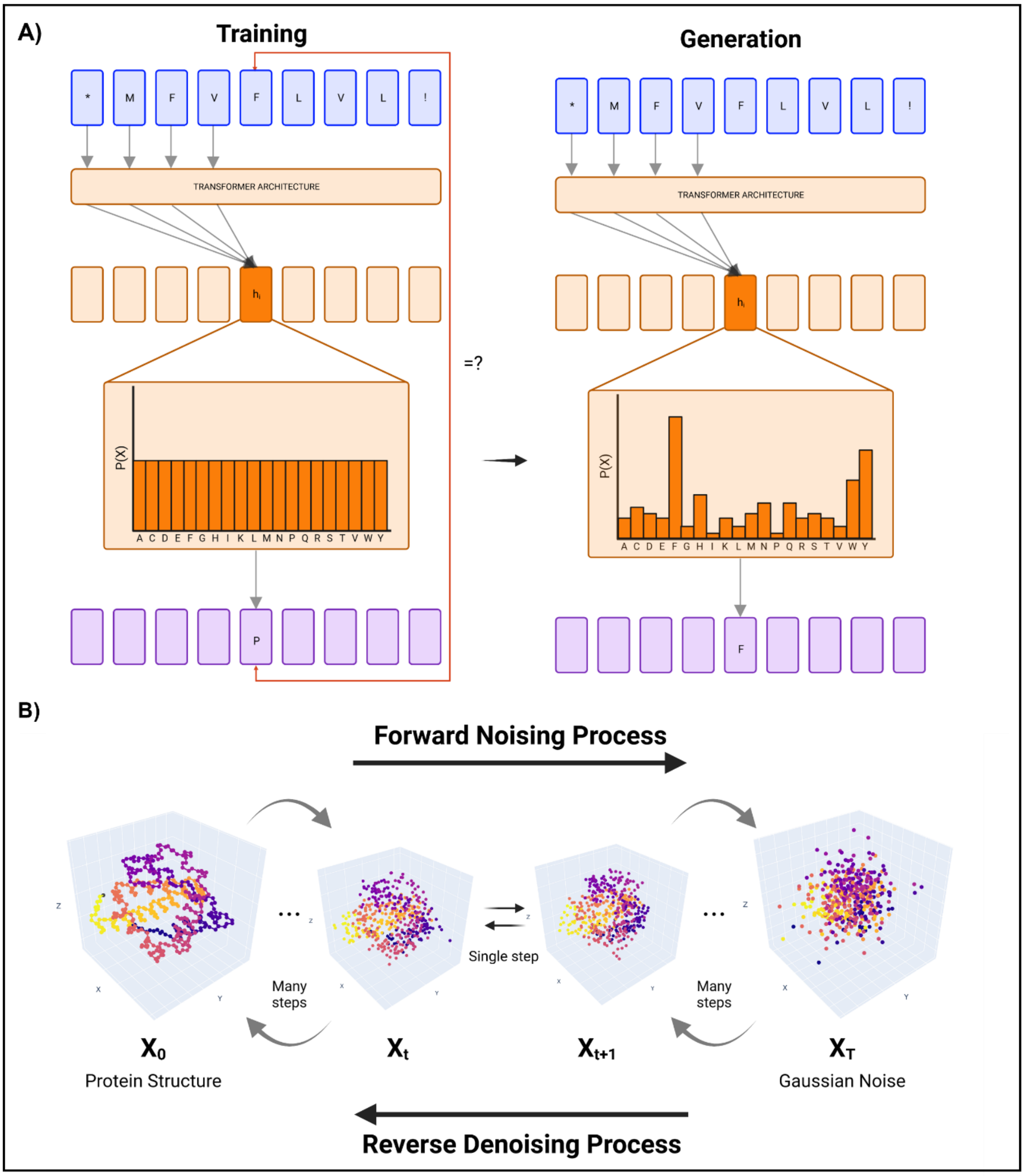

**The Training and Generation schemes for language and diffusion models a)** Language models are trained on a large sets of protein sequences without any explicit guidance on what patterns to look for. The training process involves presenting the model with a protein sequence and asking it to predict a masked amino acid in the sequence, with the exact context provided depending on the model architecture (see box 1). When the model makes a prediction, its weights are updated based on whether it made a correct or incorrect prediction (shown by the red arrow). Once trained, the model generates new protein sequences based on a given context by sampling from its learned probability distribution, which assigns probabilities to each possible next amino acid. **b)** Diffusion models are trained by taking known protein structures and noising the chosen representation of the protein (typically atomic coordinates) by applying a diffusion process which adds Gaussian noise. A neural network is then trained to learn the reverse of the noising process of known proteins (see box 2). Once trained, the model can be provided with true Gaussian noise and generate from this generate realistic protein structures. Figures adapted from (a) [44] (b) [22] and created with Biorender

**Figure 3**

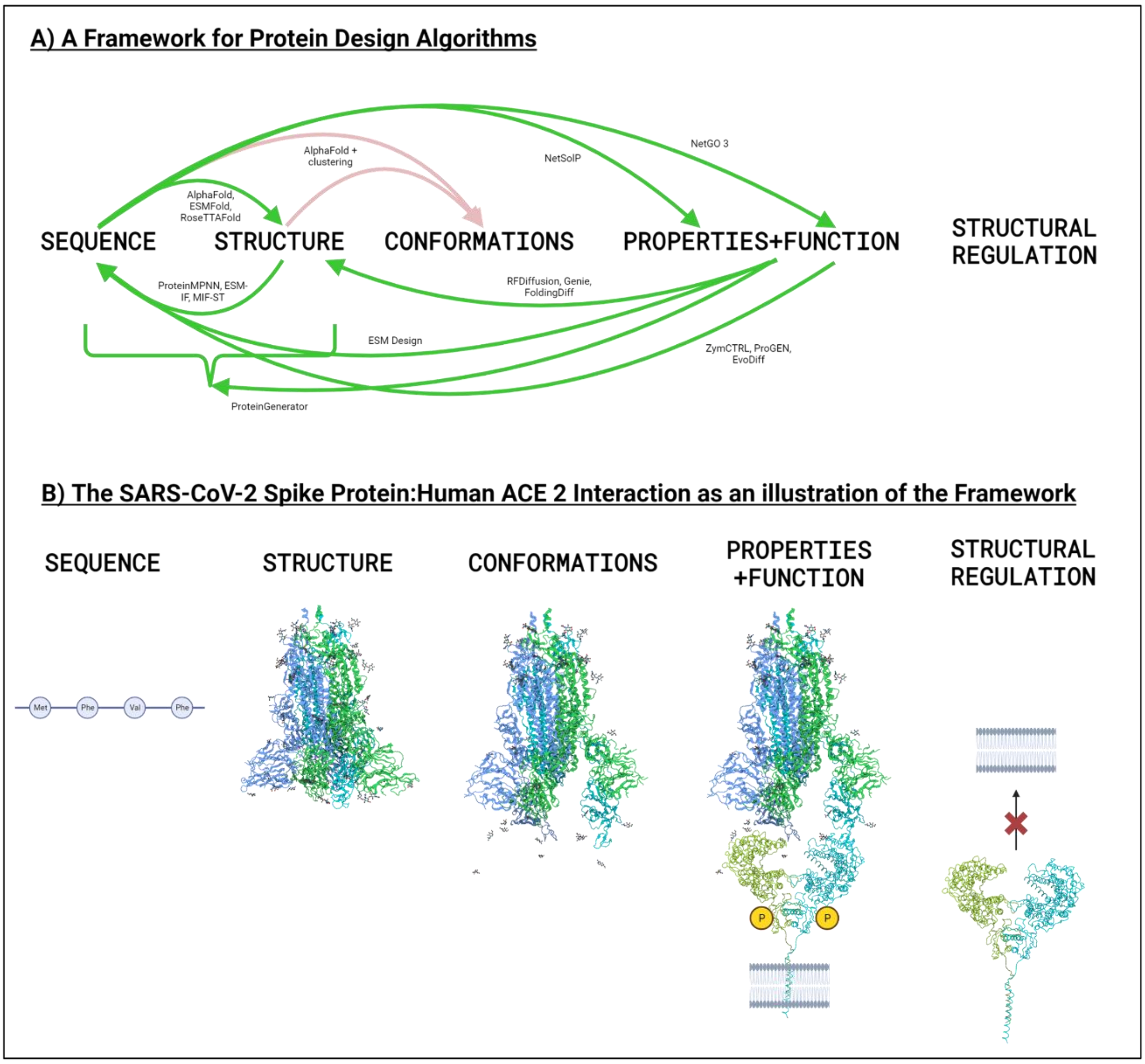

**a)** A framework to understand the inputs and outputs of computational models developed for protein design. Proteins are polymers of amino acids, forming a sequence that folds into a 3D structure. Here structure represents a single low-energy conformation, while conformations account for the entire ensemble of accessible shapes. Structural regulation represents switching on or off protein activity by addition or removal of chemical groups to the protein. Green arrows indicate rapidly advancing algorithm development. For simplicity, where there are gaps in the framework to be filled in, red arrows are used to signify tractable next steps for algorithm advancement. **b)** SARS-CoV-2 spike protein's amino acid sequence folds into a 3D structure, forming a trimer of three identical chains. The spike protein adopts different conformations, transitioning between closed and open states. The open state enables binding to human ACE2 receptors. ACE2 receptors exhibit structural regulation, with phosphate group removal reducing their cell surface presence. PDB IDs: 7DWY, 7WS8, 6M17. Figure created with Biorender.

**Box 1** Language model architectures and functional conditioning approaches. Figures adapted from (a) [44] and (b) [45]
**(a)** Large Language Models (LLMs) utilise self-supervised learning to generate learned representations of proteins that capture relationships between various amino acids. The training process involves presenting the model with a protein sequence and asking it to predict a masked amino acid in the sequence based on only the previous amino acids (autoregressive language models, e.g. GPT), all other amino acids (masked language models, e.g. BERT), or the previous and following amino acids independently (bi-directional language models) [28]. When the model makes a prediction, its weights are updated based on whether it made a correct or incorrect prediction. This process is repeated many times with the models are being trained to minimise the negative log-likelihood over the dataset $D=\{x^1,\ldots,x^{|D|}\}$. **(b)** Strategies to provide control over generated sequences' properties include: 1) high-throughput generation and filtering of sequences with desired predicted properties; 2) incorporating conditioning tags (shown); 3) employing prompt-engineering, as with RITA; and 4) fine-tuning on specific families, as in ProGEN; 5) Adopting an MCMC search approach, whereby a language model iteratively proposes mutations to an initial sequence and accepts or rejects them based on an energy function encoding desired properties and constraints. Over many iterations, sequences evolve toward low energy, meeting specified design criteria.

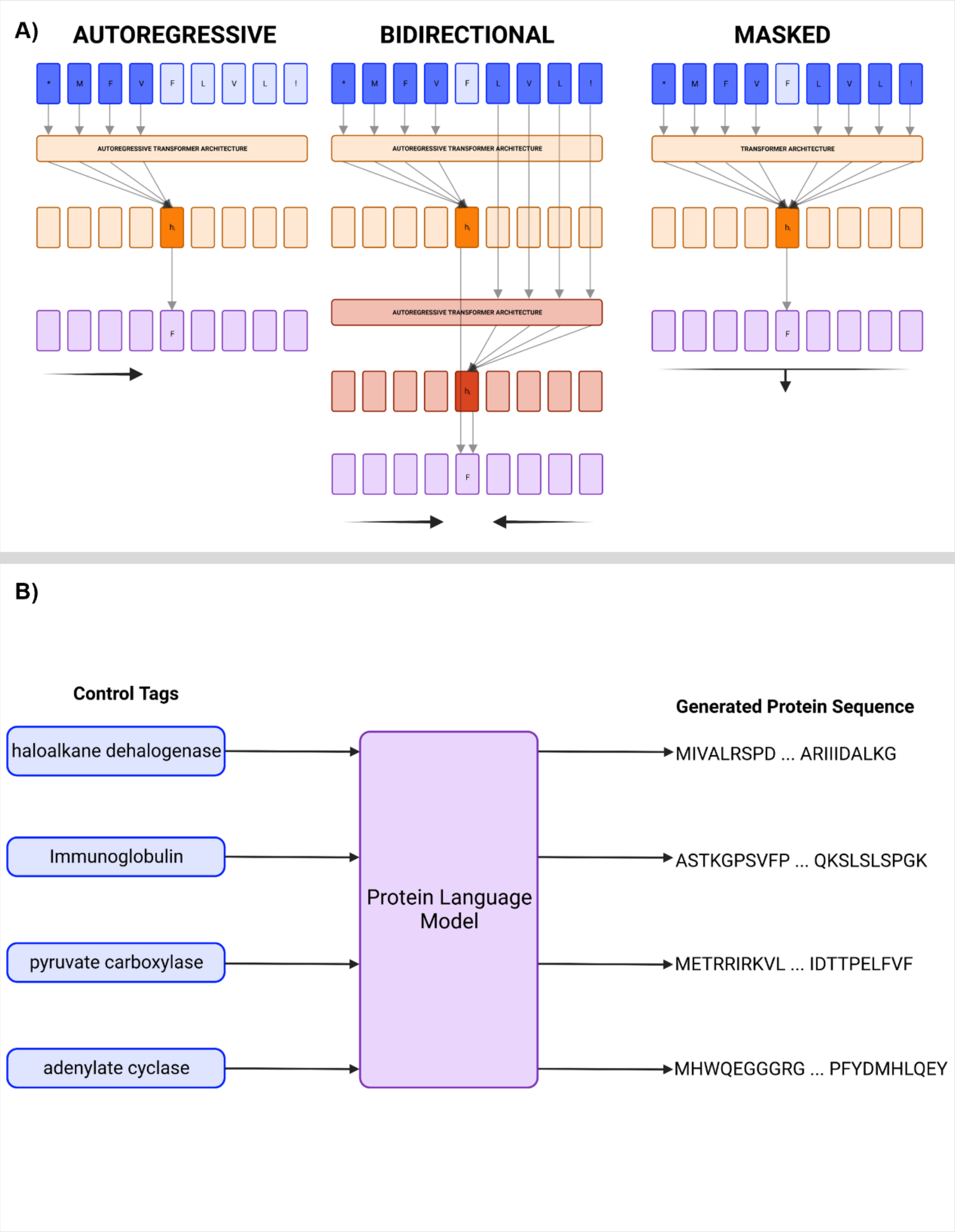

**Box 2** Diffusion model architectures and functional conditioning approaches.
Diffusion models, a class of thermodynamics-inspired generative models, have their roots in the initial concept introduced by Sohl-Dickstein in 2015. An analogous concept, known as Score Matching with Langevin Dynamics, was independently developed by Song in 2019. Ho et al 2020 built on the concept introduced by Sohl-Dickstein to implement them for generating high-quality images in a form they called Denoising Diffusion Probabilistic Models. Subsequently, these models were brought together under the umbrella of score-based generative modelling by Song in 2021. Advancements in 2022 saw Karras et al. providing a modular framework for diffusion models and Ho et al introducing progressive distillation as a novel training technique. In 2023, Song further contributed to the field by introducing consistency models. Concurrently, a unique electrostatics-inspired generative model, termed Poisson Flow Generative Models, was proposed by Xu et al. in 2022. In 2023, these authors extended Poisson Flow Generative Models into higher dimensions and integrated an augmentation dimension parameter, which surprisingly enabled the unification of these models and diffusion models under a single model family.

**A** The noise induction process of a diffusion model applied to a basic one-dimensional probability distribution function. Here, samples are drawn from a probability distribution, then Gaussian noise is progressively added by modelling a thermodynamic diffusion process. The resulting distribution closely approximates Gaussian noise.
**B** Each step in the time process generates a new noisy data distribution. As the variable 't' approaches 'T', the distribution increasingly aligns with a Gaussian.

**C** A neural network model is trained to denoised the data by predicting the reverse of the deterministic noising process across many different samples. Training starts with samples drawn from the unknown $x_0$ distribution. Using a noising trajectory of that sample, 't' is randomly sampled from a uniform distribution of [0,T]. Using Bayes rule it is possible to obtain $x_t$ directly from $x_0$. Then a neural network, known as the score function, is trained to either directly retrieve $x_0$ from $x_t$ (as per Karras et al.) or to predict the gradient of the denoising trajectory at time 't' (as per Song et al.). A loss function is then either applied to the difference between the predicted $x_0$ and the true $x_0$ or between the predicted gradient of $x_t$ and the true gradient of $x_t$. The true gradient can be obtained from the reverse of the noising process, which is deterministic and can be calculated from the forward noising process when $x_0$ is known. This training is iteratively performed over multiple timesteps, noise induction trajectories, and samples. In Denoising Diffusion Probabilistic Models (DDPMs) as introduced by Ho et al. in 2020, a different loss function of the difference between the true and predicted noise at different time steps given $x_0$ is used, which derives from a comparison of the true and predicted $x_t$ distributions.

**D** Diffusion models can be guided towards specific design objectives using Bayesian Inference during each step of the iterative generation process. To condition on a property, y, a time-dependent classifier model $p_t(y|x)$ is trained on the noised structures $x_t \sim p_t(x|x_0)$. The generative sampling process can then be simply adjusted by adding the gradient of the log likelihood that structure $x_0$ will have the desired property at time t = 0 ($\nabla x \log p_t(y|x)$) to the learned gradient from the diffusion model ($\nabla x \log p_t(x)$).

Diffusion models broadly segregate into two categories, those that incorporate stochastic elements in the denoising process (Stochastic Differential Equations, SDEs) and those that do not, relying on a specific reformulation of the reverse of the noising into an ordinary differential equation (known as the probability flow ordinary differential equation, PF ODEs). However, all these ODEs have the potential to introduce stochastic elements by adding noise (as per Karras, Chu). **The lighter shade of blue used here for the distribution of $X_T$ represents that true Gaussian noise is now being used.**
**E** During Model inference, the model is provided with true Gaussian noise. As per the training schedule, the model can either predict the gradient to guide the movement of the noised data or make a prediction of $x_0$ from the noised data and then compute the gradient based on the number of noising steps. The model then moves from the noised data to less noised data by moving the data in the direction of the predicted gradient. Further noise (ε) can then be injected at this point. In the diagram this is shown as the movement from point b to c. An alternative is to add noise before predicting $x_0$ and then compensating for the added noise by taking a larger denoising step. (a) represents the predicted of $x_0$. (b) represents the movement from $x_t$ to a less noised data distribution. (c) represents the updated $x_{t-1}$ with the injection of noise. The stochasticity can result in different generated structures from the same starting noise.
**F** The solution trajectories for many denoising tasks. Surprisingly, diffusion models generate samples from the true $x_0$ distribution that were not present in the training samples (middle trajectory).
**G** It is possible to obtain a deterministic ODE that matches the marginal evolution of the SDE. This approach results in more efficient and precise likelihood calculations ,thereby improving generation.
**H** Recent reformulations of the denoising trajectory (as per Karras et al.) set the noise step equal to time and do not scale the data, leading to more linear solution trajectories.
**I** Consistency models propose a method to directly generate denoised samples from any point on the noising trajectory.

**Box 2** Diffusion model architectures and functional conditioning approaches.

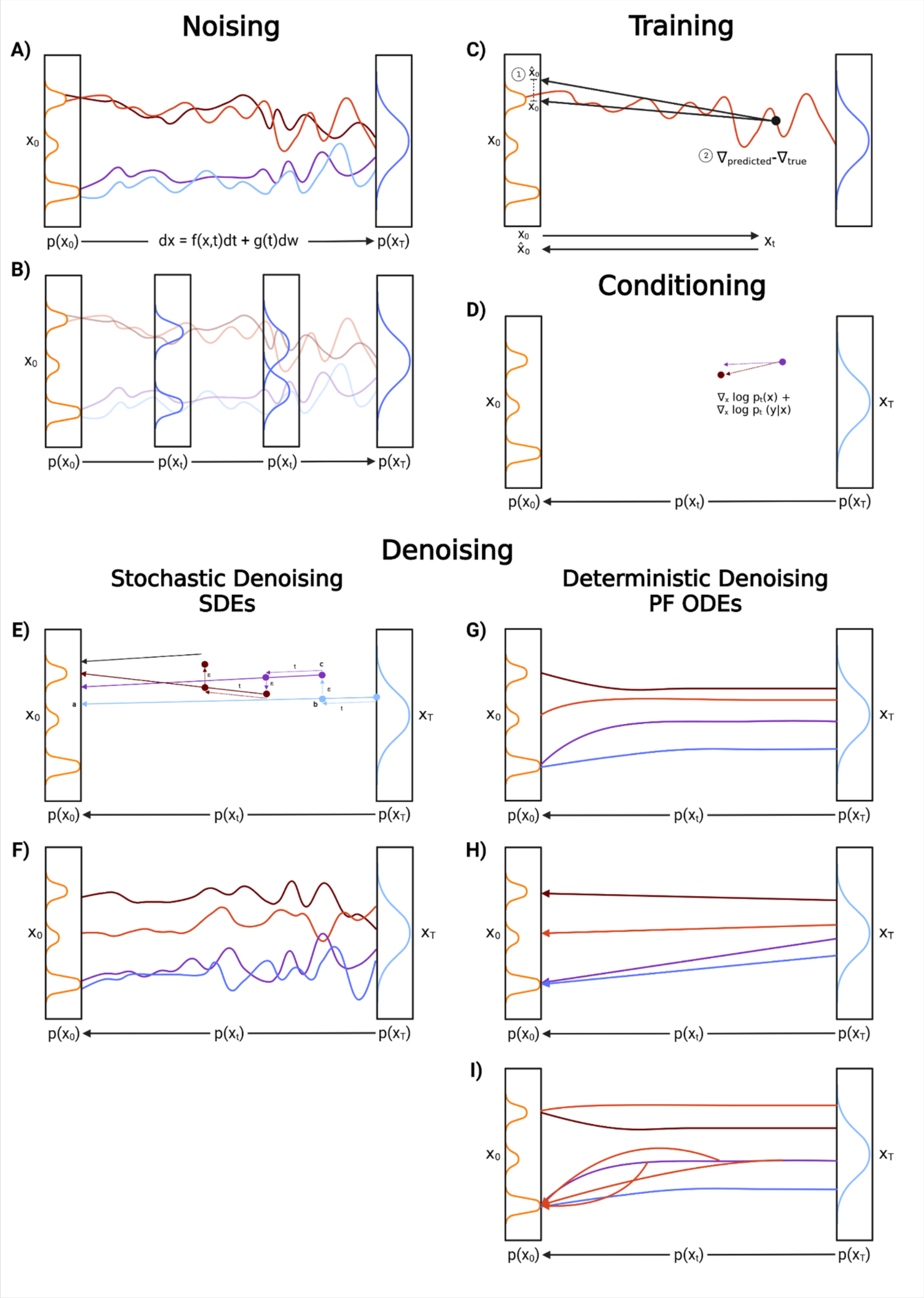

**Table 1**

**Recent generative artificial intelligence models for *de novo* protein design**

| Model Name | Ref | Controllable output | Model Type (protein representation) | Release Date (Most Recent Update) | Maximum Length | Experimental Validation | Novel Topologies | Is the code publicly available? | Is there a Google Colab tutorial or Hugging Face Space? |
|---|---|---|---|---|---|---|---|---|---|
| RITA | [37] | No | Autoregressive LLM | 01/05/2022 | N/A | No | N/A | https://github.com/lightonai/RITA | No |
| Anand Achim | [80] | Inpainting, secondary structure | Diffusion (frame representation) | 01/05/2022 | | No | Yes | No | No |
| ProtGPT2 | [34] | No | Autoregressive LLM | 01/07/2022 | 250 | No | Yes | https://huggingface.co/nferruz/ProtGPT2 | https://huggingface.co/nferruz/ProtGPT2 |
| ProteinSGM | [84] | Scaffolding | Diffusion (descriptive square matrix representation) | 01/07/2022 | 256 | Yes | Yes | https://gitlab.com/mjslee0921/proteinsgm | No |
| ProGEN2 | [35] | Enzyme family tags | Autoregressive LLM | 01/08/2022 | 512 | Yes (~100 proteins) | Yes | https://github.com/salesforce/progen | No |
| FoldingDiff | [85] | Length of desired protein | Diffusion (angle representation) | 01/09/2022 | 128 | No | N/A | https://github.com/microsoft/foldingdiff | https://huggingface.co/spaces/wukevin/foldingdiff |
| ESMDesign | [17] | Backbone shape | Masked LLM | 01/12/2022 | 184 | Yes (228 proteins) | Yes | https://github.com/facebookresearch/esm/tree/main/examples/lm-design | https://colab.research.google.com/github/facebookresearch/esm/blob/main/examples/protein-programming-language/tutorial.ipynb |
| ZymCTRL | [36] | Enzyme family tags | Autoregressive LLM | 01/12/2022 | 600 | Announced but not yet provided | Yes | https://huggingface.co/nferruz/ZymCTRL | https://huggingface.co/nferruz/ZymCTRL |
| RFDiffusion | [18] | Length, shape, scaffolding, binding partners, oligomerisation | Diffusion (frame representation) | 01/12/2022 | 600 | Yes (~1000 proteins) | Yes | https://github.com/RosettaCommons/RFdiffusion | https://colab.research.google.com/github/sokrypton/ColabDesign/blob/v1.1.1/rf/examples/diffusion.ipynb |
| Chroma | [20] | Length, shape, scaffolding, binding partners, natural language annotations, oligomerisation | Diffusion (frame representation) | 01/12/2022 | 600 | No | Yes | No | No |
| DiffSDS | [59] | Inpainting | LM + Diffusion (direction representation) | 01/01/2023 | 30 inpainted residues | No | N/A | https://github.com/A4Bio/DiffSDS_open | No |
| Genie | [82] | No | Diffusion (frame representation) | 01/02/2023 | 128 | No | Yes | https://github.com/aqlaboratory/genie | No |
| ProteinDT | [21] | Function using natural language annotations | LM + Diffusion (sequence representation) | 01/02/2023 | N/A | No | N/A | No | No |
| FrameDiff | [83] | Length | Diffusion (frame representation) | 01/05/2023 | 500 | No | Yes | https://github.com/jasonkyuyim/se3_diffusion | No |
| Protein Generator | [58] | Amino acid composition, length, scaffolding, introduction of cleavage sites | Diffusion (sequence representation) | 01/05/2023 | 300 | Yes | Yes | https://github.com/RosettaCommons/protein_generator | https://huggingface.co/spaces/merle/PROTEIN_GENERATOR |
| Protpardelle | [86] | Motif scaffolding (all atom) and inpainting | Diffusion (point cloud) | 01/05/2023 | 300 | No | Yes | https://github.com/alexechu/protpardelle | No |
| EvoDiff | [57] | MSA families, IDRs, motif scaffolding | Diffusion (sequence representation) | 12/09/2023 | N/A | No | Yes | https://github.com/microsoft/evodiff | No |
| RFDiffusionAA | [26] | Protein binders to small molecules | Diffusion (frame representation of proteins) | 09/10/2023 | 150 | Yes | Yes | Not as of publication | Not as of publication |